\documentclass[runningheads]{llncs}

\usepackage[T1]{fontenc}
\usepackage[pdftex]{graphicx}
\usepackage{amsmath}
\usepackage{bbm}
\usepackage{booktabs}
\usepackage{cite}
\usepackage{orcidlink}
\usepackage{fancyref}
\usepackage[figurename=Figure]{caption}

\begin{document}

\title{The Gentle Collapse: Distributional Metrics for Continual Learning}

\author{
        Ahmed~Anwar$^{1,2}$ \orcidlink{0009-0004-9737-7177},
        Andreas~Wagner$^{3}$ \orcidlink{0000-0001-5772-1261},
        Federico~Raue$^{1}$ \orcidlink{0000-0002-8604-6207},\\
        Tobias~Nauen$^{1,2}$ \orcidlink{0009-0007-6275-5184},
        Andreas~Dengel$^{1,2}$ \orcidlink{0000-0002-6100-8255}
}
\authorrunning{Anwar et al.: The Gentle Collapse}
\institute{
        $^{1}$German Research Center for Artificial Intelligence (DFKI), \\
        $^{2}$RPTU University Kaiserslautern-Landau, Kaiserslautern, Germany\\
        $^{3}$Hochschule Karlsruhe University of Applied Sciences, Karlsruhe, Germany
        }

\maketitle

\begin{abstract}
Accuracy degradation is the standard metric for Catastrophic Forgetting (CF), however, it records only whether forgetting occurred or not.
It saturates at the extremes and collapses discretely at task boundaries, hiding the internal structure of what is being forgotten.
We introduce six softmax-derived metrics spanning true-label rank (TLR), predictive confidence, and distributional divergence that characterize forgetting continuously, each normalized to $[0,1]$ with no modification to training.
On CIFAR-100, these metrics carry information where accuracy does not: at $0\%$ accuracy, the Confusion Margin spans an IQR of $[0.32, 0.50]$ across classes that accuracy treats identically.
We demonstrate that this richer signal is actionable in mitigating catastrophic forgetting.
Per-sample metric scores used as loss weights reduce forgetting by 1.3 percentage points over uniform experience replay (ER) on CIFAR-100.
Furthermore, the slope of a metric over a small window provides a stable sampling criterion: at a small-window size (e.g. 3 epochs), accuracy-trend degrades to $34.79\pm2.32\%$ while log-TLR achieves $41.07\pm0.57\%$.
This gap is structural since reliable small-window trend estimation requires a continuous signal.
On TinyImageNet, log-TLR trend sampling reduces forgetting by 7.7 percentage points over the ER baseline.
\keywords{Continual Learning \and Catastrophic Forgetting \and Evaluation}
\end{abstract}

\section{Introduction}

Continual learning addresses a fundamental constraint in modern machine learning: neural networks suffer from Catastrophic Forgetting (CF) of previously learned tasks when trained on new ones \cite{mccloskey1989catastrophic}.
As models grow larger and deployment settings more dynamic, this constraint becomes increasingly expensive, since full retraining is often computationally infeasible.
The field has responded with a substantial body of work on replay \cite{rolnick2019er}, regularization \cite{kirkpatrick2017ewc, zenke2017si}, and  architectural adaptation \cite{rusu2016progressive}.
Progress across all of it is measured almost universally by a single quantity: classification accuracy and its degradation across tasks.

Accuracy is a binary outcome defined at the decision boundary, however, forgetting does not happen as a binary sudden shift at the decision boundary.
It happens throughout the output distribution: confidence in the correct class erodes, the true label falls in the predicted ranking, and probability mass shifts toward newly learned categories.
When accuracy does register a drop in performance, it encodes no information about whether a class is marginally forgotten or entirely displaced.
Our experiments validate this: in our CIFAR-100 experiment, in 31 observations where the class accuracy is exactly $0\%$ shown in Figure \ref{fig:conditional_violins}, the Confusion Margin spans an IQR of $[0.32, 0.50]$ and True-Label Rank a median of $0.65$, both metrics continuing to distinguish degrees of forgetting that accuracy collapses to a single value.
Two models that accuracy declares indistinguishable may differ substantially in how much residual knowledge they retain and how amenable that knowledge is to recovery.

We introduce six metrics derived from the softmax output distribution that characterize forgetting at a finer resolution than accuracy allows.
They fall into three families: true-label rank metrics (TLR, Log-TLR, Reciprocal-TLR), which track the position of the correct class in the predicted ranking as forgetting progresses; confidence and confusion metrics (True-Label Confidence, Confusion Margin), which quantify predictive certainty and proximity to misclassification; and a distribution-matching metric (Normalized KL), which measures global output-distribution shift relative to the ideal one-hot encoded target.
All six are normalized to $[0,1]$, interpretable alongside accuracy, and require no additional forward passes or architectural changes.

The argument of this paper is three-part.
First, the metrics expose distributional structure that accuracy cannot see: they carry information at the extremes where accuracy saturates and change gradually where accuracy steps.
Second, this structure is actionable at the sample level, since metric scores assign a meaningful gradient of forgetting severity to individual replay samples, and using them to re-weight loss contributions improves performance over uniform replay.
Third, this structure is actionable at the temporal level: the slope of a metric over a short trailing window is a stable and informative signal for replay prioritization.
At a window of three epochs on CIFAR-100, accuracy-trend sampling collapses to $34.79\pm2.32\%$ while log-TLR trend sampling achieves $41.07\pm0.57\%$, a gap that reflects the structural impossibility of reliable short-window trend estimation on a binary signal.
Continuous distributional metrics do not share this limitation.

Our contributions are as follows.
We introduce six normalized, softmax-derived metrics for continual learning evaluation and show empirically that they reveal forgetting
dynamics invisible to accuracy-based analysis.
We show that per-sample metric scores are effective loss weights for replay, with consistent results on CIFAR-100 and TinyImageNet.
We show that metric trend over a short window is an effective and stable sampling criterion, that accuracy-trend sampling exhibits catastrophic instability at short windows that continuous metrics do not, and that this asymmetry is structural not quantitative.
Taken together, these results argue that the field's reliance on accuracy has not only limited evaluation; it has limited what interventions are possible for mitigating CF.

\section{Related Work}

\textbf{Continual learning methods.}
Catastrophic forgetting \cite{mccloskey1989catastrophic} is addressed by
three families of methods \cite{wang2024comprehensive,delange2021survey}.
Regularization-based methods protect past knowledge by penalising
changes to important parameters (EWC \cite{kirkpatrick2017ewc},
SI \cite{zenke2017si}) or by self-distillation on new-task data
(LwF \cite{li2017lwf}).
Replay-based methods store a buffer of past examples and interleave them
with current training.
Standard ER \cite{rolnick2019er} uses uniform sampling; GEM
\cite{lopez2017gem} and A-GEM \cite{chaudhry2019agem} add gradient
constraints; iCaRL \cite{rebuffi2017icarl} pairs replay with a
herding-based exemplar selector; DER/DER++ \cite{buzzega2020der} replays
stored logits alongside inputs, combining cross-entropy with a
logit-matching distillation term to form a strong baseline; and X-DER
\cite{boschini2022xder} further revises stored logits across tasks.
Architecture-based methods grow separate capacity per task
\cite{rusu2016progressive} or isolate sub-networks \cite{serra2018hat},
avoiding interference at the cost of scalability.

\textbf{Evaluation of continual learning.}
CL evaluation has long centred on the accuracy matrix
$A = \{a_{k,j}\}$ \cite{lopez2017gem}, from which forward/backward
transfer and forgetting measures are derived.
D\'{i}az-Rodr\'{i}guez et al.\ \cite{diazrodriguez2018donot} were among
the first to challenge this, arguing that memory overhead, compute cost,
and positive transfer deserve equal standing, and proposing a composite
CLScore.
The CLEVA-Compass \cite{mundt2022cleva} formalised multi-dimensional
evaluation as a visual framework promoting transparency across diverse
scenarios.
Despite these efforts, all introduced frameworks still tie catastrophic
forgetting to the binary classification accuracy signal.
Most relevant to our work, De Lange et al.\ \cite{delange2023stability}
showed through per-iteration evaluation that all major CL methods suffer
from a stability gap, a sharp transient performance drop immediately
after a new task begins, which end-of-task evaluation misses entirely.
Their metrics quantify worst-case accuracy; ours quantify the degree of
forgetting continuously through the full output distribution, and further
show that this richer signal enables principled intervention at both the
sample and temporal level.

\textbf{Output-based metrics in classification.}
Our metrics are all derived from the softmax distribution rather than
the top-1 decision.
True Label Confidence and Normalized KL are grounded in calibration
research \cite{guo2017calibration}, where the probability assigned to
the true class is recognised as a proper scoring signal independent of
the hard prediction.
The TLR family extends rank-based evaluation from information retrieval
(MRR, nDCG \cite{manning2008ir}) and top-$k$ accuracy
\cite{russakovsky2015imagenet} into a continuous scalar in $[0,1]$,
tracking how far the true label is pushed down the ranked list as
forgetting progresses.
The Confusion Margin is related to inference-time decision margins
\cite{liu2016margin}, but used here purely as an evaluation signal
rather than a training objective.

\section{Methodology}
\label{sec:methodology}

Standard continual learning evaluation centres on the accuracy matrix $A = \{a_{k,j}\}$ \cite{wang2024comprehensive,mundt2022cleva}, where $a_{k,j}\in[0,1]$ is the accuracy of task $j$ after learning task $k$.
Aggregate statistics such as average accuracy, forgetting measure, and backward/forward transfer are all derived from this matrix \cite{lopez2017gem}.
Because every entry $a_{k,j}$ is a hard decision-boundary outcome, all of these measures inherit the same coarseness: two models that produce identical accuracy matrices can differ substantially in how much residual knowledge they retain.

To see why this matters, consider a classifier trained on four classes ($0$--$3$) in two tasks.
After task 2, models $A$ and $B$ both misclassify a test sample with true label $y=0$, giving $a_{2,1}=0$ in both cases.
Yet their softmax outputs tell a very different story:
\[
  f_2^A(x) = [\textbf{0.48},\;0.49,\;0.02,\;0.01],
  \qquad
  f_2^B(x) = [\textbf{0.05},\;0.10,\;0.70,\;0.15].
\]
Model $A$ has nearly recovered the true class, while model $B$ has almost entirely forgotten it, a distinction that no accuracy-based measure can express.

\subsection{Proposed Metrics}
\label{subsec:metrics}

We focus on a classification problem with $C$ classes learned over $T$ tasks.
Let $f_k(x_i^j)[c]$ denote the $c$-th softmax output of model $f_k$, after training on task $k$, on sample $x_i^j$ from task $j$, and let $y_i^j$ be its true label.
All six metrics below are defined on a test set $D_j$ of size $n_j$, take values in $[0,1]$, and are oriented so that higher is better, matching the convention of accuracy.

\subsubsection{Confusion Margin (CM).}
The margin between the top predicted probability and the probability assigned to the true label directly measures proximity to misclassification:
\[
  m_{k,j} = \frac{1}{n_j}\sum_{i=1}^{n_j}
    \Bigl(1 + f_k(x_i^j)[y_i^j] - \max_c f_k(x_i^j)[c]\Bigr).
\]
A value of 1 corresponds to a perfectly confident correct prediction; lower values indicate increasing confusion.

\subsubsection{True-Label Rank family (TLR, LTLR, RTLR).}
These metrics track the position of the true class in the ranked softmax
output, where rank 1 denotes a correct classification.
Formally,
$\mathrm{rank}(y_i^j, f_k(x_i))
  = 1 + \sum_{c \neq y_i^j}
    \mathbbm{1}(f_k(x_i)[c] > f_k(x_i)[y_i^j])$.
Rank-based evaluation extends established information-retrieval practice
(MRR, nDCG \cite{manning2008ir}) and top-$k$ accuracy
\cite{russakovsky2015imagenet} into a continuous scalar.
The three variants trade off sensitivity at different parts of the rank
distribution:
\begin{align*}
r_{k,j} &= \frac{1}{n_j}\sum_{i=1}^{n_j}
  \left(1 - \frac{\mathrm{rank}(y_i^j,f_k(x_i^j))-1}{C-1}\right),\\[4pt]
l_{k,j} &= \frac{1}{n_j}\sum_{i=1}^{n_j}
  \left(1 - \frac{\ln\bigl(\mathrm{rank}(y_i^j,f_k(x_i^j))\bigr)}
                 {\ln C}\right),\\[4pt]
s_{k,j} &= \frac{1}{n_j}\sum_{i=1}^{n_j}
  \frac{C - \mathrm{rank}(y_i^j,f_k(x_i^j))}
       {(C-1)\cdot\mathrm{rank}(y_i^j,f_k(x_i^j))}.
\end{align*}
TLR ($r$) penalises all rank displacements equally (linearly); Log-TLR ($l$) applies a logarithmic compression that softens penalties for distant errors; Reciprocal-TLR ($s$) concentrates sensitivity at the top ranks, making it the strictest measure for near-correct predictions.

\subsubsection{True Label Confidence (CTL).}
Mean predicted probability of the true class is the simplest distributional signal, directly grounded in calibration research \cite{guo2017calibration}:
\[
  q_{k,j} = \frac{1}{n_j}\sum_{i=1}^{n_j} f_k(x_i^j)[y_i^j].
\]

\subsubsection{Normalized KL (NKL).}
The KL divergence from the model's output to the Dirac distribution on the true class equals $-\ln f_k(x_i)[y_i]$, confirming that CTL corresponds to $1-\exp(\mathrm{KL})$.
For a complementary normalization that caps divergence at the uniform-distribution baseline $\ln C$, we define:
\[
  k_{k,j} = \frac{1}{n_j}\sum_{i=1}^{n_j}
    \left(1 - \frac{\min\bigl(-\ln f_k(x_i^j)[y_i^j],\;\ln C\bigr)}
                   {\ln C}\right).
\]
NKL captures global output-distribution shift and is the most sensitive metric to severe forgetting where the model assigns near-zero probability to the true class.



\section{Experiments}
\label{sec:experiments}

\subsection{Experimental Setup}
\label{subsec:setup}

Diagnostic experiments (Section \ref{subsec:dynamics}) use Sequential CIFAR-100 with a 5-task class-incremental split (20 classes per task).
We train a ResNet-18 backbone with a replay buffer of size 2000, batch size 256, learning rate 0.03, and 50 epochs per task, using Experience Replay (ER) with uniform sampling as the base method.
Metrics are computed per class at every epoch, enabling fine-grained analysis across the full training trajectory.
Intervention experiments (Sections \ref{subsec:weighted_loss}--\ref{subsec:trend_sampling}) additionally include TinyImageNet (200 classes, 10 tasks) to assess generalisation to a harder benchmark.
All intervention results are reported as mean $\pm$ std across three independent runs.
The window size $w=10$ was selected based on the sensitivity analysis in Section \ref{subsec:trend_sampling}.

\subsection{Metrics Reveal Forgetting Dynamics}
\label{subsec:dynamics}

\subsubsection{A gradual signal where accuracy is binary.}
\begin{figure}[t]
  \centering
  \includegraphics[width=\linewidth]{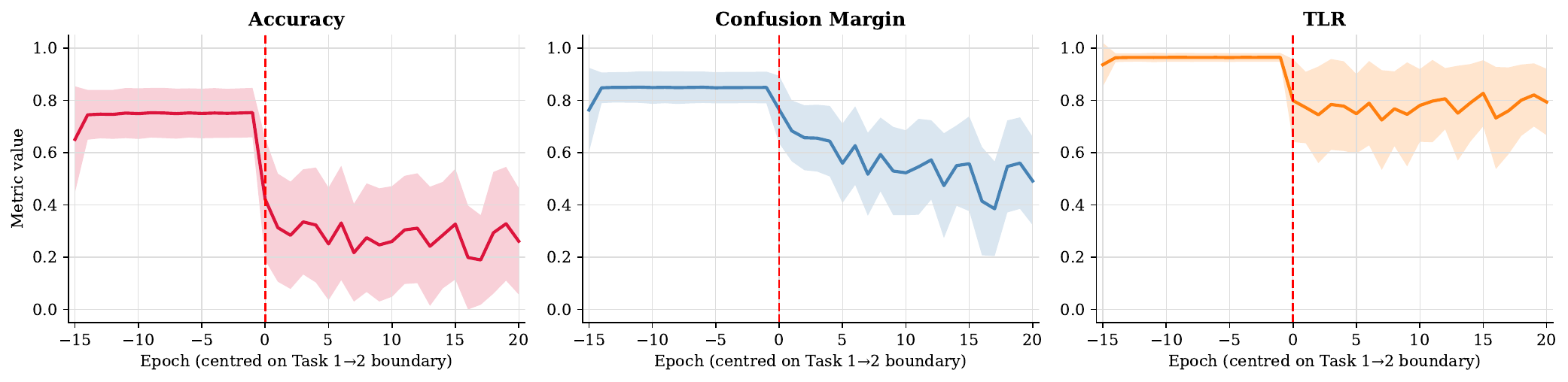}
  \caption{Mean $\pm$ std across Task-1 classes for accuracy, confusion
           margin, and true-label rank centred on the Task 1$\to$2
           boundary (red dashed line).
           Continuous metrics decay gradually while accuracy collapses
           sharply, exposing a graded forgetting signal that accuracy
           cannot provide.}
  \label{fig:trajectories}
\end{figure}

\Fref{fig:trajectories} shows mean $\pm$ std across Task-1 classes for accuracy, CM, and TLR centred on the Task 1$\to$2 boundary.
When the new task begins, accuracy collapses sharply, dropping to near zero within a few epochs with no intermediate signal.
The continuous metrics respond to the same event differently: they decay gradually over many  epochs, exposing a continuum of forgetting severity that the binary accuracy signal compresses into a single step.
At any point after the task boundary, the continuous metrics still carry information about how severely a class has been forgotten, while accuracy has already saturated for most classes.

\begin{figure}[t]
  \centering
  \includegraphics[trim={550 0 0 90},clip, scale=0.4]{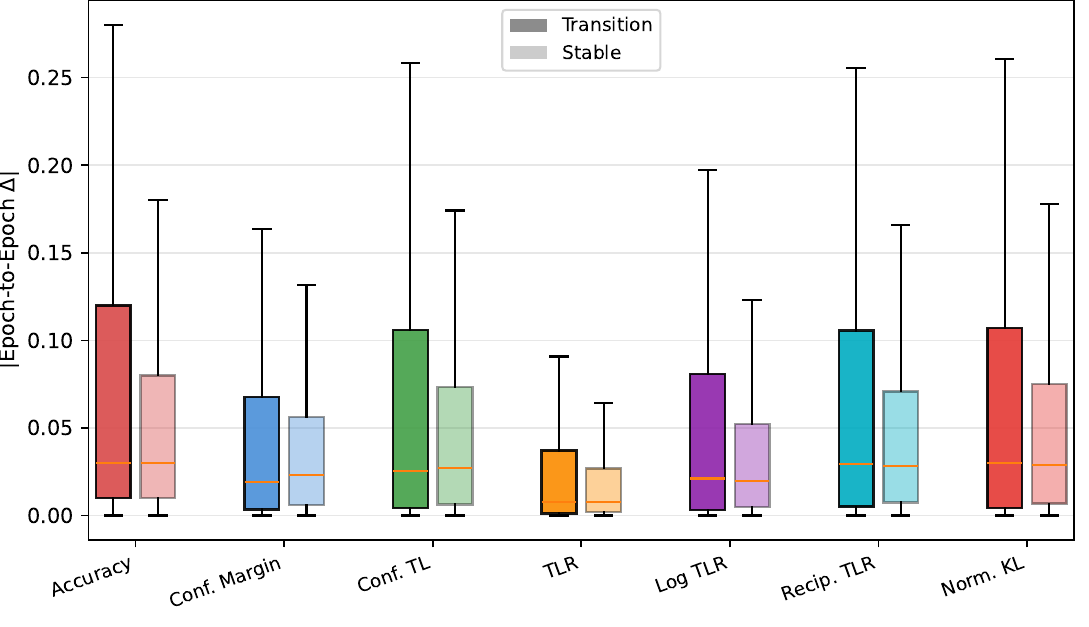}
  \caption{Epoch-to-epoch absolute change split by phase: transition
           epochs ($\pm5$ steps around each task boundary) versus stable
           mid-task epochs.
           Accuracy produces extreme jumps at transitions (upper
           whisker $>0.35$).
           CTL, RTLR, and NKL are more volatile compared to CM, TLR, and Log-TLR.}
  \label{fig:smoothness}
\end{figure}

\Fref{fig:smoothness} quantifies this by separating epochs into transition epochs (within $\pm5$ steps of a task boundary) and stable epochs (mid-task).
Accuracy produces extreme outlier jumps at boundaries, with the upper whisker exceeding 0.35, and these spikes are absent for all continuous metrics.
The continuous metrics show a moderate and consistent elevation at transitions.
When a task boundary occurs, accuracy undergoes a structural discontinuity; the continuous metrics register a moderate, trackable increase in change rate.
This smoothness is not merely diagnostic: it is precisely what makes short-window trend estimation tractable, as we show in Section \ref{subsec:trend_sampling}.

\subsubsection{Information where accuracy is saturated.}
\begin{figure}[t]
  \centering
  \includegraphics[trim={0 0 0 8},clip, scale=0.5]{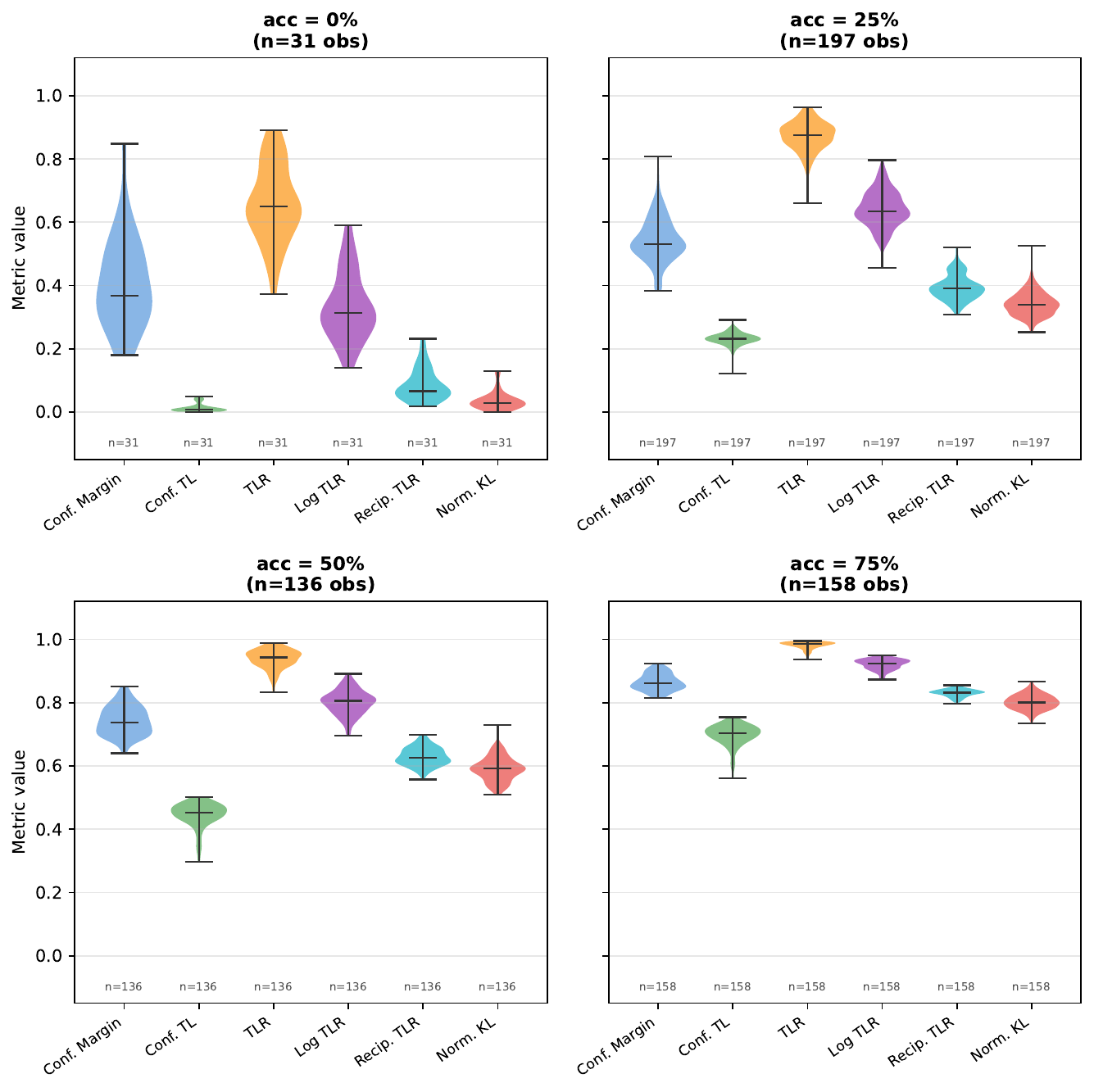}
  \caption{Distribution of each metric at fixed accuracy levels
           ($\pm0.5\%$ windows).
           At $0\%$ accuracy, CM and TLR retain wide distributions
           while CTL and NKL are more concentrated.
           Across all intermediate levels, the metrics carry substantial
           variation beyond what accuracy encodes.}
  \label{fig:conditional_violins}
\end{figure}

When accuracy reaches zero for a class, it provides no further signal.
Figure \ref{fig:conditional_violins} shows the distribution of each metric restricted to the $n$ epochs where class-accuracy is in the corresponding accuracy bin $\pm0.5\%$.
At bin $acc=0\%$ (when accuracy reaches zero) all samples are indistinguishable accuracy-wise, however CM spans an IQR of $[0.32, 0.50]$ with values reaching 0.85, and TLR has a median of 0.65 with IQR $[0.60, 0.76]$, which gives these metrics a discriminative power that can be exploited.
A class with $\mathrm{CM}=0.85$ has its true label nearly at the top of the predicted ranking despite being classified incorrectly; a class with $\mathrm{CM}=0.18$ has been pushed far from the decision boundary.
Accuracy reports both identically nevertheless, which further highlights the drawbacks of relying on accuracy alone in understanding CF.

CTL and NKL show very little spread at $0\%$ accuracy; IQR $[0.00, 0.01]$ and $[0.02, 0.04]$ respectively: at zero accuracy, the model systematically assigns low probability to the true class, so confidence-based measures saturate near zero alongside accuracy.
This pattern holds across fixed accuracy levels from $0\%$ to $75\%$: rank-based and margin-based metrics remain informative at the extremes where confidence-based measures do not as shown in Figure \ref{fig:conditional_violins}.
The complementarity of rank-based and confidence-based metrics guides metric selection for the intervention experiments that follow: NKL and CTL, which carry strong signal in mid-range forgetting, are suited to sample-level weighting; TLR and Log-TLR, which remain discriminative at severe forgetting, are the more robust choice for trend estimation on harder tasks.

\begin{table}[t]
\centering
\caption{Metric-weighted loss on Sequential CIFAR-100 and TinyImageNet
         ($n=3$ seeds). Higher accuracy and lower forgetting are better.
         }
\label{tab:weighted_loss}
\begin{tabular}{lcccc}
\toprule
& \multicolumn{2}{c}{\textbf{CIFAR-100}} & \multicolumn{2}{c}{\textbf{TinyImageNet}} \\
\cmidrule(lr){2-3}\cmidrule(lr){4-5}
\textbf{Weight Metric} & \textbf{Acc.\ (\%)}$\uparrow$ & \textbf{Forg.\ (\%) $\downarrow$}
                       & \textbf{Acc.\ (\%)}$\uparrow$ & \textbf{Forg.\ (\%)}$\downarrow$ \\
\midrule
ER uniform (baseline) & $36.33\pm0.55$ & $53.20\pm0.59$
                      & $23.67\pm0.81$ & $54.98\pm0.28$ \\
\midrule
TLR            & $36.40\pm0.68$ & $52.30\pm0.31$
               & $23.51\pm0.38$ & $55.00\pm1.06$ \\
Log-TLR        & $\mathbf{36.98\pm0.55}$ & $51.92\pm0.35$
               & $23.72\pm0.19$ & $\mathbf{53.38\pm1.58}$ \\
Reciprocal-TLR & $35.42\pm0.49$ & $52.55\pm0.49$
               & $23.04\pm0.30$ & $54.67\pm0.12$ \\
Confusion Margin & $36.43\pm0.40$ & $52.54\pm0.16$
                 & $\mathbf{23.76\pm0.60}$ & $54.05\pm1.08$ \\
Normalized KL  & $35.56\pm0.55$ & $\mathbf{51.49\pm0.22}$
               & $23.30\pm0.46$ & $53.85\pm0.12$ \\
Confidence TL  & $35.42\pm0.54$ & $52.10\pm0.51$
               & $22.78\pm0.90$ & $53.89\pm0.72$ \\
\midrule
Accuracy       & $35.21\pm0.58$ & $52.21\pm0.84$
                   & $22.19\pm0.87$ & $54.54\pm0.60$ \\
\bottomrule
\end{tabular}
\end{table}
\subsection{Metric-Weighted Loss}
\label{subsec:weighted_loss}

The distributional metrics assign a continuous score to each sample reflecting the current severity of forgetting for its class.
We investigate whether concentrating gradient contributions on more-forgotten samples, by using per-sample metric scores as loss weights, produces measurable gains over uniform ER.

At each replay step, each sample is assigned a weight inversely proportional to its current metric score, so that more-forgotten samples exert a larger influence on the parameter update.
Table \ref{tab:weighted_loss} reports final accuracy and forgetting on Sequential CIFAR-100 and TinyImageNet for each metric-weighted variant.
On CIFAR-100, Log-TLR produces the highest accuracy ($36.98\pm0.55\%$) and a forgetting reduction of 1.3 percentage points over the ER baseline ($51.92\pm0.35\%$ vs $53.20\pm0.59\%$).
Confusion Margin and TLR perform similarly ($36.43\pm0.40\%$ and $36.40\pm0.68\%$ respectively), while confidence-based metrics trail by roughly 1 percentage point.
On TinyImageNet, accuracy improvements are within noise across all variants, but the forgetting reduction is consistent: Log-TLR reduces forgetting from $54.98\pm0.28\%$ to $53.38\pm1.58\%$, and NKL achieves $53.85\pm0.12\%$ with lower variance.
Accuracy-weighted loss ($35.21\pm0.58\%$ on CIFAR-100, $22.19\pm0.87\%$ on TinyImageNet) serves as a confirmation that sample reweighting helps in principle,  but being a binary signal it fails to outperform the baseline, while all continuous distributional variants do.


\begin{table}[t]
\centering
\caption{Metric trend sampling on Sequential CIFAR-100 and TinyImageNet
         ($w=10$, $\gamma=2.0$, $n=3$ seeds).
         Bold marks the best result per column.}
\label{tab:trend_tinyimagenet}
\begin{tabular}{lcccc}
\toprule
& \multicolumn{2}{c}{\textbf{CIFAR-100}} & \multicolumn{2}{c}{\textbf{TinyImageNet}} \\
\cmidrule(lr){2-3}\cmidrule(lr){4-5}
\textbf{Trend metric} & \textbf{Acc.\ (\%)}$\uparrow$ & \textbf{Forg.\ (\%)$\downarrow$}
                      & \textbf{Acc.\ (\%)}$\uparrow$ & \textbf{Forg.\ (\%)}$\downarrow$ \\
\midrule
ER uniform (baseline) & $36.33\pm0.55$ & $53.20\pm0.59$
                      & $23.67\pm0.81$ & $54.98\pm0.28$ \\
\midrule
TLR            & $39.93\pm0.94$          & $46.10\pm1.99$
               & $23.14\pm0.72$          & $46.79\pm1.21$ \\
Log-TLR        & $41.71\pm0.73$          & $44.52\pm0.57$
               & $\mathbf{24.74\pm0.81}$ & $47.26\pm1.14$ \\
Reciprocal-TLR & $\mathbf{41.88\pm0.60}$ & $44.73\pm1.07$
               & $24.16\pm1.34$          & $49.11\pm1.37$ \\
Confusion Margin & $40.67\pm1.14$        & $46.30\pm1.31$
                 & $23.27\pm0.13$        & $49.39\pm3.34$ \\
Normalized KL  & $41.13\pm0.70$          & $45.95\pm0.77$
               & $21.92\pm4.23$          & $46.54\pm3.95$ \\
Confidence TL  & $37.30\pm1.34$  &  $51.21\pm2.02$  
               & $23.43\pm1.15$ & $48.32\pm0.91$ \\  
\midrule
Loss           & $41.44\pm0.52$ & $\mathbf{43.73\pm0.52}$
               & $24.00\pm1.37$ & $\mathbf{46.23\pm3.53}$ \\
Accuracy       & $37.75\pm0.94$ & $50.65\pm1.35$
               & $24.30\pm0.27$ & $49.11\pm1.74$ \\
\bottomrule
\end{tabular}
\end{table}

\subsection{Metric Trend Sampling}
\label{subsec:trend_sampling}

Sample-level metric scores capture the current degree of forgetting for each class.
A complementary question is whether the rate of change of a metric over a recent window provides additional information for replay prioritization.
The sampling weight is computed as $\text{softmax}(\gamma \cdot slope)$ over classes, where $\gamma = 2.0$ controls the sharpness of prioritization and slope is estimated by linear regression over the trailing window of $w$ epochs.
A class whose metric score is declining rapidly is actively being forgotten; a class whose score is stable may not benefit as urgently from additional replay even if its current score is low.

\subsubsection{Window sensitivity on CIFAR-100.}
\begin{figure}[t]
  \centering
  \includegraphics[scale=0.50]{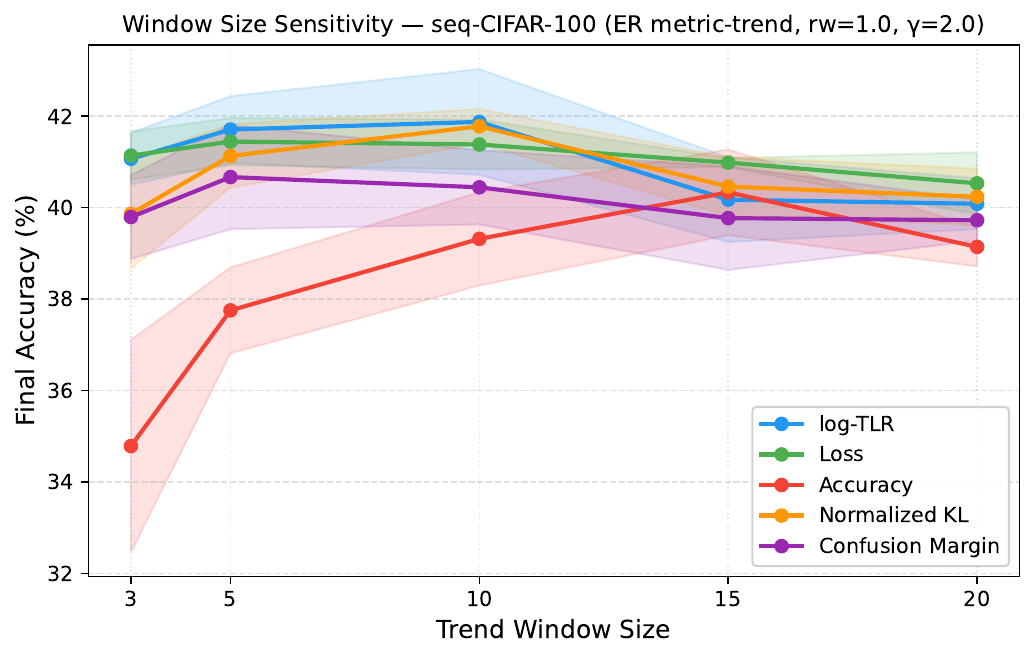}
  \caption{Final accuracy on Sequential CIFAR-100 as a function of
           trend window size ($\gamma=2.0$, $n=3$ seeds, shaded regions
           show $\pm1$ std).
           At $w=3$, accuracy-trend sampling collapses to
           $34.79\pm2.32\%$, more than six percentage points below
           every continuous metric at the same window.
           All continuous signals are stable across all window sizes.
           The instability of accuracy-trend is structural: slope
           estimation on a binary signal is dominated by the noise of
           discrete threshold crossings.}
  \label{fig:window_sweep}
\end{figure}

Figure \ref{fig:window_sweep} shows final accuracy on Sequential CIFAR-100 as a  function of window size for five trend signals: Log-TLR, Loss, Normalized KL, Confusion Margin, and Accuracy.

At $w=3$, accuracy-trend sampling collapses to $34.79\pm2.32\%$, with error bands reaching below $33\%$.
Every continuous signal, Log-TLR, Loss, Normalized KL, and Confusion Margin, is stable at $w=3$, clustering between $39.8\%$ and $41.1\%$.
As the window grows, accuracy partially recovers, reaching $40.33\%$ at $w=15$ before falling to $39.14\%$ at $w=20$, and its variance remains elevated throughout.

Accuracy at the class level is a step function; it moves in discrete jumps as samples cross the decision boundary, and the slope of a step function over a short window is dominated by the noise of those transitions.
A longer history is required to average that noise out.
This is exactly what Figure \ref{fig:window_sweep} shows: accuracy needs $w \geq 15$ before approaching the performance that continuous metrics achieve at $w=3$.

The key distinction is between continuous and binary signals, not between our proposed metrics and all alternatives.
Loss is also a continuous signal and is correspondingly stable across windows ($41.14\pm0.53\%$ at $w=3$ through $40.53\pm0.68\%$ at $w=20$).
Table \ref{tab:trend_tinyimagenet} shows that at $w=10$, Loss achieves the best forgetting on CIFAR-100 ($43.73\pm0.52\%$), while Reciprocal-TLR leads on accuracy ($41.88\pm0.60\%$); the continuous metrics are broadly competitive, and all substantially outperform accuracy-trend.
In settings where the trend window cannot be tuned, rank-based metrics are the safer default, given their additional stability on harder tasks
(see below).

Furthermore, Table \ref{tab:trend_tinyimagenet} shows results on TinyImageNet at $w=10$.
Log-TLR achieves a 7.7 percentage point reduction in forgetting relative to the ER baseline ($47.26\pm1.14\%$ vs $54.98\pm0.28\%$), with a corresponding improvement in final accuracy and low variance across seeds.
Loss-trend matches log-TLR on forgetting mean ($46.23\%$) but with considerably higher variance ($\pm3.53\%$ vs $\pm1.14\%$), suggesting that rank-based metrics provide a more stable trend signal on harder tasks.
Accuracy-trend is competitive on final accuracy ($24.30\pm0.27\%$) but substantially weaker on forgetting ($49.11\pm1.74\%$), reflecting the partial stabilisation of the binary signal at $w=10$ noted in the window analysis.

Normalized KL and CM show elevated variance ($\pm4.23\%$ and $\pm3.34\%$).
While CM performed well in metric-weighted loss, its spread at 0\% accuracy visible in Figure \ref{fig:conditional_violins} results in sensitivity in a metric trend scenario.
Normalized KL, like loss and CTL, is less robust under trend estimation than rank-based metrics, likely because its sensitivity to low-probability tails amplifies noise on the harder task.

\section{Conclusion}
\label{sec:conclusion}

Accuracy reduces model behaviour to a binary signal at the decision boundary, it is therefore blind to the gradual distributional shifts that precede and accompany forgetting. 
The lack of this nuance hinders its application to fine-grained interventions that could mitigate CF.

We introduced six softmax-derived metrics spanning true-label rank, confidence and confusion, and distribution matching.
On Sequential CIFAR-100, these metrics expose structure that accuracy cannot: they carry substantial information where accuracy is saturated, distinguishing degrees of forgetting that accuracy collapses to a single value, and they evolve continuously at task boundaries where accuracy undergoes discontinuities.
We showed that this richer characterisation is actionable in two ways.
Per-sample metric scores used as loss weights reduce forgetting by 1.3 percentage points over uniform ER on CIFAR-100, with consistent forgetting improvements on TinyImageNet despite modest accuracy gains.
Metric slope over a short trailing window reduces forgetting on TinyImageNet by 7.7 percentage points and improves accuracy under log-TLR.
The temporal actionability depends on the signals being continuous: at a window of three epochs, accuracy-trend sampling collapses to $34.79\pm2.32\%$ while log-TLR maintains $41.07\pm0.57\%$, a gap that reflects the structural impossibility of reliable short-window trend estimation on a binary signal.

The metrics impose no additional training cost and are complementary to existing evaluation frameworks.
More broadly, the granularity of an evaluation signal determines the granularity of the interventions it can support.
Extending metric-guided interventions to stronger replay methods such as DER++\cite{buzzega2020der} and iCaRL\cite{rebuffi2017icarl}, where the interaction between logit-matching distillation and distributional metric-weighting is non-trivial, is a natural direction for future work. 

\begin{credits}
\subsubsection{\ackname}
This work was supported by the BMFTR project Albatross (funding code 16IW24002).
All compute was done thanks to the Pegasus cluster at DFKI Kaiserslautern.
\subsubsection{\discintname}
The authors have no competing interests to declare that are relevant to
the content of this article.
\end{credits}

\bibliographystyle{splncs04}
\bibliography{references}

\end{document}